\pdfoutput=1

\documentclass[final,12pt]{elsarticle}

\usepackage{etoolbox}
\makeatletter
\patchcmd{\ps@pprintTitle}{\footnotesize\itshape
       Preprint submitted to \ifx\@journal\@empty Elsevier
       \else\@journal\fi\hfill\today}{\relax}{}{}
\makeatother

\usepackage{hyperref}
\usepackage{ccicons}

\usepackage{amssymb}

\journal{Physics of Life Reviews}

\begin{document}

\begin{frontmatter}


\ead{carlos.gomez@udc.es}
\ead[url]{http://www.grupolys.org/~cgomezr}


\title{On the relation between dependency distance, crossing dependencies, and parsing.\\Comment on ``Dependency distance: a new perspective on syntactic patterns in natural languages'' by Haitao Liu et al.}


\author{Carlos G\'{o}mez-Rodr\'{i}guez}

\address{Universidade da Coru\~{n}a. FASTPARSE Lab, LyS Research Group, Departamento de Computaci\'{o}n, Facultade de Inform\'{a}tica, Elvi\~{n}a, 15071 A Coru\~{n}a, Spain}



\begin{keyword}
dependency distance \sep natural language parsing \sep dependency parsing \sep crossing dependencies \sep non-projectivity



\end{keyword}

\end{frontmatter}

\vspace{1cm}
\noindent\fbox{%
    \parbox{\textwidth}{%
        $^{\scriptsize{\textcopyright}}$ 2017. This manuscript version is made available under the CC-BY-NC-ND 4.0 license (\url{http://creativecommons.org/licenses/by-nc-nd/4.0/}). \ccbyncnd\\ This is the accepted manuscript (final peer-reviewed manuscript) accepted for publication in Physics of Life Reviews, and may not reflect subsequent changes resulting from the publishing process such as copy-editing, formatting or pagination. The published journal article can be found at \url{https://doi.org/10.1016/j.plrev.2017.05.007}.
    }%
}
\vspace{1cm}





Liu et al. \cite{Liu2017} provide a comprehensive account of research on dependency distance in human languages. 
While the article is a very rich and useful report on this complex subject, here I will expand on a few specific issues where research in computational linguistics (specifically natural language processing) can inform DDM research, and vice versa. These aspects have not been explored much in \cite{Liu2017} or elsewhere, probably due to the little overlap between both research communities, 
but they may provide interesting insights for 
improving our understanding of the evolution of human languages, the mechanisms by which the brain processes and understands language, and the construction of effective computer systems to achieve this goal.

\section{Crossings, dependency distance and the parallelism between exceptions to projectivity and exceptions to DDM}

As mentioned in \cite{Liu2017}, there is a close relation between DDM and the scarcity of crossing dependencies in natural languages. This low frequency of crossings has long been observed \cite{melcuk88}, later quantified \cite{Hav07,GomNivCL2013,GomCL2016}, and recently statistically tested \cite{FerGomEstArxiv2017}, in a wide range of human languages.

It is worth noting, however, that strict projectivity (a prohibition of crossing dependencies) is not an adequate model of the syntax of real sentences. One the one hand, it fails to explain a number of relevant linguistic phenomena present in various languages \cite{Levy2012a,Versley2014}. On the other hand, an overwhelming majority of syntactic corpora in recent multilingual collections have been observed to contain non-projectivity \cite{GomNivCL2013,GomCL2016}: crossing dependencies are scarce, but far from absent \cite{FerGomEstArxiv2017}. For these reasons, while projectivity can be useful from an engineering point of view, in the context of a tradeoff between coverage and efficiency in natural language parsers \cite{GomCL2016}; taking it for granted when investigating DDM or using it as an assumption of models to explain DDM \cite{Liu2008a,Futrell2015a} can lead to a methodological pitfall: the scarcity of crossing dependencies is likely not an independent constraint of language that contributes to DDM, but rather a consequence of it \cite{GomFerArxiv2016,FerGomComplexity2016}.

To account for the limited amount of crossing dependencies that arise in language and to be able to build parsers that can handle non-projective syntactic phenomena in an efficient manner, researchers have explored various classes of so-called mildly non-projective dependency structures \cite{KuhNiv06,GomCarWeiCL2011,Pitler2012,GomNivCL2013,Pitler2013,GomCL2016}. These are sets of trees that allow a limited degree of non-projectivity, permitting crossing dependencies only if they follow certain conditions or patterns. Various such classes have been defined that claim very high coverage over the trees in a variety of corpora, allowing a large majority of the non-projective dependencies that appear in practice. However, the reasons for this success (and especially, the reasons why some of the proposed sets have more coverage than others) are currently not very well known. Namely, the reasons for the high coverage of each given mildly non-projective class could include either being an adequate description of the particular situations where crossing dependencies can arise, or yielding a large enough class of syntactic trees to provide high coverage by sheer brute force, or a mix of both. The observation that a given class can be more or less adequate depending on the criteria used to annotate the syntactic dependencies \cite{GomCL2016} suggests at least some influence of the first factor.

The research on syntactic patterns involving long-distance dependencies, reviewed by Liu et al. in Section 5 of \cite{Liu2017}, could help clarify this question. DDM and the scarcity of crossing dependencies are closely related, as the latter is motivated by short dependency distance \cite{FerGomComplexity2016}. Furthermore, in both cases, there is a predominant trend (the majority of dependencies are short and do not cross) but there are exceptions that escape the trend (long-distance dependencies and crossing dependencies), which are in turn related (longer dependencies are more likely to cross \cite{Ferrer2014f}). Liu et al. \cite{Liu2017} review some possible reasons for the minority of long dependencies observed in language, and explanations of how they can survive the pressure for DDM. This raises two questions: (1) do these explanations also apply to the presence of crossing dependencies, and are they related with the adequacy of mildly non-projective classes of trees? (e.g., do the most effective such classes work well because they favor crossing dependencies from words at peripheral positions, or structures that can be easily chunked?); and (2) can we in turn re-use some of what we know about long dependencies for crossing dependencies and their parsing, and employ it to define classes of mildly non-projective structures that will more closely adjust to the kinds of non-projectivity found in language? Both questions are interesting avenues for research, and can advance our knowledge both on DDM and natural language parsing.

\section{The surprising effectiveness of transition-based parsers and their bias towards DDM}

Liu et al. \cite{Liu2017} cite some work in computational linguistics that achieved improvements in parsing accuracy by purposefully introducing dependency distance as a constraint in a parser \cite{DBLP:books/daglib/p/EisnerS10}. Additionally, it is worth noting that many state-of-the-art natural language parsers use algorithms with an implicit bias towards short dependency distances, even if they do not introduce it as an explicit restriction.

In particular, a popular framework for dependency parsers is the transition-based (or shift-reduce) approach \cite{Nivre2008}, under which a parser is defined with a non-deterministic state machine, a statistical or machine-learning-based model to score transitions, and a search strategy to obtain the optimal sequence of transitions that will yield a parse. Many, if not most, of the current state-of-the-art parsing systems are based on this framework \cite{CheMan2014,dyer-EtAl:2015:ACL-IJCNLP,andor-EtAl:2016:P16-1,TACL885,Alberti17}, and all the different algorithm variants that are at its core have in common that they build short dependencies before (and requiring fewer transitions than) long ones, be it because building a long dependency requires removing intervening nodes from a stack (as in the popular arc-standard and arc-eager \cite{Nivre2008}, arc-hybrid \cite{KuhGomSatACL2011} or swap \cite{nivre09acl} algorithms) or because it requires to navigate a list (as in the systems based on the Covington \cite{covington01} algorithm). This bias towards favoring short dependencies can be part of the reason why these systems are so effective in practice, and is an example of the trend pointed out in \cite{GomBBS2016} by which purely engineering-oriented parsing models are converging with cognitive theories of language understanding, even when they do not have psycholinguistic modeling among their goals.

A quick verification and quantification of the mentioned bias can be undertaken by implementing transition systems and obtaining random trees by taking a random transition at each state. Focusing on sentences of length 20 as an example, the expected mean dependency distance for a uniformly random tree is 7 \cite{Ferrer2004b}, contrasting with real averages observed in corpora, e.g. 2.52 for Arabic, 3.18 for German, 2.59 for English or 2.51 for Spanish in the sentences of length 20 of the Stanford HamleDT 2.0 treebanks \cite{HamledTStanford}. On the other hand, by simulating $10^5$ random parses of sentences of length $20$, we obtain an average distance of 2.44 with the arc-standard algorithm, and 2.06 with the arc-eager parser with the tree constraint \cite{NivreAEP14}. These two parsers are projective, but using an algorithm that can generate arbitrary non-projective trees (the swap parser), the obtained average is 2.38, even smaller than for arc-standard. While further investigation is needed that would be outside the scope of this comment, the data seem to suggest that transition-based parsers implicitly favor dependency lengths that are equal, or even smaller, than the natural ones that appear in language as a consequence of DDM; and this could be an important factor in the practical adequacy of these parsers. 

\section*{Acknowledgements}
 
This research has received funding from the European Research Council (ERC) under the European Union's Horizon 2020 research and innovation programme (grant agreement No 714150 - FASTPARSE), and from the TELEPARES-UDC project (FFI2014-51978-C2-2-R) from MINECO. I thank Ramon Ferrer-i-Cancho for helpful comments.



\bibliographystyle{elsarticle-num} 
\bibliography{twoplanaracl,carlos-own,main,Ramon}

\newcommand{\beeksort}[1]{}
\begin{thebibliography}{10}
\expandafter\ifx\csname url\endcsname\relax
  \def\url#1{\texttt{#1}}\fi
\expandafter\ifx\csname urlprefix\endcsname\relax\def\urlprefix{URL }\fi
\expandafter\ifx\csname href\endcsname\relax
  \def\href#1#2{#2} \def\path#1{#1}\fi

\bibitem{Liu2017}
H.~Liu, C.~Xu, J.~Liang,
  \href{http://www.sciencedirect.com/science/article/pii/S1571064517300532}{Dependency
  distance: A new perspective on syntactic patterns in natural languages},
  Physics of Life Reviews (in this issue).
\newblock \href
  {http://dx.doi.org/http://dx.doi.org/10.1016/j.plrev.2017.03.002}
  {\path{doi:http://dx.doi.org/10.1016/j.plrev.2017.03.002}}.
\newline\urlprefix\url{http://www.sciencedirect.com/science/article/pii/S1571064517300532}

\bibitem{melcuk88}
I.~Mel'{\v{c}}uk, Dependency Syntax: Theory and Practice, State University of
  New York Press, 1988.

\bibitem{Hav07}
J.~Havelka, Beyond projectivity: Multilingual evaluation of constraints and
  measures on non-projective structures, in: ACL 2007: Proceedings of the 45th
  Annual Meeting of the Association for Computational Linguistics, 2007, pp.
  608--615.

\bibitem{GomNivCL2013}
C.~G\'{o}mez-Rodr\'{\i}guez, J.~Nivre,
  \href{http://dx.doi.org/10.1162/COLI\_a\_00150}{Divisible transition systems
  and multiplanar dependency parsing}, Comput. Linguist. 39~(4) (2013)
  799--845.
\newblock \href {http://dx.doi.org/10.1162/COLI\_a\_00150}
  {\path{doi:10.1162/COLI\_a\_00150}}.
\newline\urlprefix\url{http://dx.doi.org/10.1162/COLI\_a\_00150}

\bibitem{GomCL2016}
C.~G\'{o}mez-Rodr\'{i}guez,
  \href{http://dx.doi.org/10.1162/COLI\_a\_00267}{Restricted non-projectivity:
  Coverage vs. efficiency}, Computational Linguistics 42~(4) (2016) 809--817.
\newblock \href {http://dx.doi.org/10.1162/COLI\_a\_00267}
  {\path{doi:10.1162/COLI\_a\_00267}}.
\newline\urlprefix\url{http://dx.doi.org/10.1162/COLI\_a\_00267}

\bibitem{FerGomEstArxiv2017}
R.~{Ferrer-i-Cancho}, C.~G\'omez-Rodr\'iguez, J.~L. Esteban,
  \href{https://arxiv.org/abs/1703.08324}{Are crossing dependencies really
  scarce?}, arXiv 1703.08324 [physics.soc-ph].
\newline\urlprefix\url{https://arxiv.org/abs/1703.08324}

\bibitem{Levy2012a}
R.~Levy, E.~Fedorenko, M.~Breen, T.~Gibson, The processing of extraposed
  structures in {English}, Cognition 122~(1) (2012) 12 -- 36.

\bibitem{Versley2014}
Y.~Versley, \href{http://www.aclweb.org/anthology/W14-6104}{Experiments with
  easy-first nonprojective constituent parsing}, in: Proceedings of the First
  Joint Workshop on Statistical Parsing of Morphologically Rich Languages and
  Syntactic Analysis of Non-Canonical Languages, Dublin City University,
  Dublin, Ireland, 2014, pp. 39--53.
\newline\urlprefix\url{http://www.aclweb.org/anthology/W14-6104}

\bibitem{Liu2008a}
H.~Liu, Dependency distance as a metric of language comprehension difficulty,
  Journal of Cognitive Science 9 (2008) 159--191.

\bibitem{Futrell2015a}
R.~Futrell, K.~Mahowald, E.~Gibson,
  \href{http://www.pnas.org/content/112/33/10336.abstract}{Large-scale evidence
  of dependency length minimization in 37 languages}, Proceedings of the
  National Academy of Sciences 112~(33) (2015) 10336--10341.
\newblock \href
  {http://arxiv.org/abs/http://www.pnas.org/content/112/33/10336.full.pdf}
  {\path{arXiv:http://www.pnas.org/content/112/33/10336.full.pdf}}, \href
  {http://dx.doi.org/10.1073/pnas.1502134112}
  {\path{doi:10.1073/pnas.1502134112}}.
\newline\urlprefix\url{http://www.pnas.org/content/112/33/10336.abstract}

\bibitem{GomFerArxiv2016}
C.~G\'omez-Rodr\'iguez, R.~{Ferrer-i-Cancho},
  \href{http://arxiv.org/abs/1601.03210}{The scarcity of crossing dependencies:
  a direct outcome of a specific constraint?}, arXiv 1601.03210 [cs.CL].
\newline\urlprefix\url{http://arxiv.org/abs/1601.03210}

\bibitem{FerGomComplexity2016}
R.~{Ferrer-i-Cancho}, C.~G\'{o}mez-Rodr\'{i}guez,
  \href{http://dx.doi.org/10.1002/cplx.21810}{Crossings as a side effect of
  dependency lengths}, Complexity 21~(S2) (2016) 320--328.
\newblock \href {http://dx.doi.org/10.1002/cplx.21810}
  {\path{doi:10.1002/cplx.21810}}.
\newline\urlprefix\url{http://dx.doi.org/10.1002/cplx.21810}

\bibitem{KuhNiv06}
M.~Kuhlmann, J.~Nivre, Mildly non-projective dependency structures, in:
  Proceedings of the COLING/ACL 2006 Main Conference Poster Sessions, 2006, pp.
  507--514.

\bibitem{GomCarWeiCL2011}
C.~G{\'{o}}mez{-}Rodr{\'{\i}}guez, J.~A. Carroll, D.~J. Weir,
  \href{http://dx.doi.org/10.1162/COLI\_a\_00060}{Dependency parsing schemata
  and mildly non-projective dependency parsing}, Computational Linguistics
  37~(3) (2011) 541--586.
\newblock \href {http://dx.doi.org/10.1162/COLI\_a\_00060}
  {\path{doi:10.1162/COLI\_a\_00060}}.
\newline\urlprefix\url{http://dx.doi.org/10.1162/COLI\_a\_00060}

\bibitem{Pitler2012}
E.~Pitler, S.~Kannan, M.~Marcus,
  \href{http://www.aclweb.org/anthology/D12-1044}{Dynamic programming for
  higher order parsing of gap-minding trees}, in: Proceedings of the 2012 Joint
  Conference on Empirical Methods in Natural Language Processing and
  Computational Natural Language Learning, Association for Computational
  Linguistics, Jeju Island, Korea, 2012, pp. 478--488.
\newline\urlprefix\url{http://www.aclweb.org/anthology/D12-1044}

\bibitem{Pitler2013}
E.~Pitler, S.~Kannan, M.~Marcus,
  \href{http://aclweb.org/anthology/Q13-1002}{Finding optimal
  1-endpoint-crossing trees}, Transactions of the Association of Computational
  Linguistics 1 (2013) 13--24.
\newline\urlprefix\url{http://aclweb.org/anthology/Q13-1002}

\bibitem{Ferrer2014f}
R.~{Ferrer-i-Cancho}, Non-crossing dependencies: least effort, not grammar, in:
  A.~Mehler, A.~L{\"u}cking, S.~Banisch, P.~Blanchard, B.~Job (Eds.), Towards a
  theoretical framework for analyzing complex linguistic networks, Springer,
  Berlin, 2016, pp. 203--234.

\bibitem{DBLP:books/daglib/p/EisnerS10}
J.~Eisner, N.~A. Smith,
  \href{http://dx.doi.org/10.1007/978-90-481-9352-3\_8}{Favor short
  dependencies: Parsing with soft and hard constraints on dependency length},
  in: H.~Bunt, P.~Merlo, J.~Nivre (Eds.), Trends in Parsing Technology,
  Dependency Parsing, Domain Adaptation, and Deep Parsing, Springer, 2010, pp.
  121--150.
\newblock \href {http://dx.doi.org/10.1007/978-90-481-9352-3\_8}
  {\path{doi:10.1007/978-90-481-9352-3\_8}}.
\newline\urlprefix\url{http://dx.doi.org/10.1007/978-90-481-9352-3\_8}

\bibitem{Nivre2008}
J.~Nivre,
  \href{http://www.mitpressjournals.org/doi/abs/10.1162/coli.07-056-R1-07-027}{{Algorithms
  for Deterministic Incremental Dependency Parsing}}, Computational Linguistics
  34~(4) (2008) 513--553.
\newblock \href {http://dx.doi.org/10.1162/coli.07-056-R1-07-027}
  {\path{doi:10.1162/coli.07-056-R1-07-027}}.
\newline\urlprefix\url{http://www.mitpressjournals.org/doi/abs/10.1162/coli.07-056-R1-07-027}

\bibitem{CheMan2014}
D.~Chen, C.~Manning, \href{http://www.aclweb.org/anthology/D14-1082}{A fast and
  accurate dependency parser using neural networks}, in: Proceedings of the
  2014 Conference on Empirical Methods in Natural Language Processing (EMNLP),
  Doha, Qatar, 2014, pp. 740--750.
\newline\urlprefix\url{http://www.aclweb.org/anthology/D14-1082}

\bibitem{dyer-EtAl:2015:ACL-IJCNLP}
C.~Dyer, M.~Ballesteros, W.~Ling, A.~Matthews, N.~A. Smith,
  \href{http://www.aclweb.org/anthology/P15-1033}{Transition-based dependency
  parsing with stack long short-term memory}, in: Proceedings of the 53rd
  Annual Meeting of the Association for Computational Linguistics and the 7th
  International Joint Conference on Natural Language Processing (Volume 1: Long
  Papers), Association for Computational Linguistics, Beijing, China, 2015, pp.
  334--343.
\newline\urlprefix\url{http://www.aclweb.org/anthology/P15-1033}

\bibitem{andor-EtAl:2016:P16-1}
D.~Andor, C.~Alberti, D.~Weiss, A.~Severyn, A.~Presta, K.~Ganchev, S.~Petrov,
  M.~Collins, \href{http://www.aclweb.org/anthology/P16-1231}{Globally
  normalized transition-based neural networks}, in: Proceedings of the 54th
  Annual Meeting of the Association for Computational Linguistics (Volume 1:
  Long Papers), Association for Computational Linguistics, Berlin, Germany,
  2016, pp. 2442--2452.
\newline\urlprefix\url{http://www.aclweb.org/anthology/P16-1231}

\bibitem{TACL885}
E.~Kiperwasser, Y.~Goldberg,
  \href{https://transacl.org/ojs/index.php/tacl/article/view/885}{Simple and
  accurate dependency parsing using bidirectional lstm feature
  representations}, Transactions of the Association for Computational
  Linguistics 4 (2016) 313--327.
\newline\urlprefix\url{https://transacl.org/ojs/index.php/tacl/article/view/885}

\bibitem{Alberti17}
C.~Alberti, D.~Andor, I.~Bogatyy, M.~Collins, D.~Gillick, L.~Kong, T.~Koo,
  J.~Ma, M.~Omernick, S.~Petrov, C.~Thanapirom, Z.~Tung, D.~Weiss,
  \href{http://arxiv.org/abs/1703.04929}{Syntaxnet models for the conll 2017
  shared task}, CoRR abs/1703.04929.
\newline\urlprefix\url{http://arxiv.org/abs/1703.04929}

\bibitem{KuhGomSatACL2011}
M.~Kuhlmann, C.~G\'{o}mez-Rodr\'{i}guez, G.~Satta,
  \href{http://www.aclweb.org/anthology/P11-1068}{Dynamic programming
  algorithms for transition-based dependency parsers}, in: Proceedings of the
  49th Annual Meeting of the Association for Computational Linguistics: Human
  Language Technologies (ACL 2011), Association for Computational Linguistics,
  Portland, Oregon, USA, 2011, pp. 673--682.
\newline\urlprefix\url{http://www.aclweb.org/anthology/P11-1068}

\bibitem{nivre09acl}
J.~Nivre, Non-projective dependency parsing in expected linear time, in:
  Proceedings of the Joint Conference of the 47th Annual Meeting of the ACL and
  the 4th International Joint Conference on Natural Language Processing of the
  AFNLP (ACL-IJCNLP), 2009, pp. 351--359.

\bibitem{covington01}
M.~A. Covington, A fundamental algorithm for dependency parsing, in:
  Proceedings of the 39th Annual ACM Southeast Conference, 2001, pp. 95--102.

\bibitem{GomBBS2016}
C.~G\'omez-Rodr\'iguez,
  \href{http://journals.cambridge.org/article\_S0140525X15000795}{Natural
  language processing and the {Now-or-Never} bottleneck}, Behavioral and Brain
  Sciences 39 (2016) e74.
\newblock \href {http://dx.doi.org/10.1017/S0140525X15000795}
  {\path{doi:10.1017/S0140525X15000795}}.
\newline\urlprefix\url{http://journals.cambridge.org/article\_S0140525X15000795}

\bibitem{Ferrer2004b}
R.~{Ferrer-i-Cancho}, {Euclidean} distance between syntactically linked words,
  Physical Review E 70 (2004) 056135.

\bibitem{HamledTStanford}
R.~Rosa, J.~Ma{\v{s}}ek, D.~Mare{\v{c}}ek, M.~Popel, D.~Zeman,
  Z.~{\v{Z}}abokrtsk{\'{y}}, {HamleDT} 2.0: Thirty dependency treebanks
  stanfordized, in: N.~C.~C. Chair), K.~Choukri, T.~Declerck, H.~Loftsson,
  B.~Maegaard, J.~Mariani, A.~Moreno, J.~Odijk, S.~Piperidis (Eds.),
  Proceedings of the Ninth International Conference on Language Resources and
  Evaluation (LREC'14), European Language Resources Association (ELRA),
  Reykjavik, Iceland, 2014.

\bibitem{NivreAEP14}
J.~Nivre, D.~Fern{\'a}ndez-Gonz{\'a}lez,
  \href{http://www.aclweb.org/anthology/J/J14/J14-2002.pdf}{Arc-eager parsing
  with the tree constraint}, Computational Linguistics 40~(2) (2014) 259--267.
\newline\urlprefix\url{http://www.aclweb.org/anthology/J/J14/J14-2002.pdf}

\end{thebibliography}






\end{document}